\documentclass[sigconf, nonacm]{acmart}
\usepackage{enumitem}
\usepackage{booktabs}
\usepackage{adjustbox}
\usepackage[ruled,linesnumbered]{algorithm2e}
\SetKw{Continue}{continue}
\usepackage{algorithmicx}
\usepackage{adjustbox}
\usepackage{multirow}
\usepackage{tabularx} 
\usepackage{array}
\newcolumntype{Y}{>{\centering\arraybackslash}X}

\newcommand\vldbdoi{XX.XX/XXX.XX}
\newcommand\vldbpages{XXX-XXX}
\newcommand\vldbvolume{14}
\newcommand\vldbissue{1}
\newcommand\vldbyear{2020}
\newcommand{\vpara}[1]{\vspace{0.05in}\noindent \textbf{#1 }}
\newcommand{\ipara}[1]{\vspace{0.05in}\noindent \textit{#1 }}
\newcommand{\brain}{Central Agent }
\newcommand{\model}{\textsc{ACTS-SQL} }
\newcommand\vldbauthors{\authors}
\newcommand\vldbtitle{\shorttitle} 
\newcommand\vldbavailabilityurl{URL_TO_YOUR_ARTIFACTS}
\newcommand\vldbpagestyle{plain} 

\begin{document}
\title{ACTS-SQL: Agentic and Critic-Oriented Tree-Structured SQL Correctness with Large Language Models}


\author{Xinmei Huang}
\affiliation{%
  \institution{Renmin University of China}
}
\email{huangxinmei@ruc.edu.cn}

\author{Jie Song}
\affiliation{%
  \institution{ByteDance Inc.}
}
\email{jie.song@bytedance.com}

\author{Peng Li}
\affiliation{%
  \institution{ByteDance Inc.}
}
\email{peng.li01@bytedance.com}

\author{Fuxin Jiang}
\affiliation{%
  \institution{ByteDance Inc.}
}
\email{jiangfuxin@bytedance.com}

\author{Jing Zhang}
\affiliation{%
  \institution{Renmin University of China}
}
\email{zhang-jing@ruc.edu.cn}

\author{Tieying Zhang}
\affiliation{%
  \institution{ByteDance Inc.}
}
\email{tieying.zhang@bytedance.com}

\author{Jianjun Chen}
\affiliation{%
  \institution{ByteDance Inc.}
}
\email{jianjun.chen@bytedance.com}

\author{Chenming Liu}
\affiliation{%
  \institution{ByteDance Inc.}
}
\email{liuchenming.123@bytedance.com}

\author{Tao Yang}
\affiliation{%
  \institution{ByteDance Inc.}
}
\email{yangtao.alan@bytedance.com}

\author{Maoyin Liu}
\affiliation{%
  \institution{ByteDance Inc.}
}
\email{liumaoyin@bytedance.com}

\author{Wenda Li}
\affiliation{%
  \institution{ByteDance Inc.}
}
\email{liwenda.wonder@bytedance.com}

\author{Hong Chen}
\affiliation{%
  \institution{Renmin University of China}
}
\email{chong@ruc.edu.cn}

\author{Cuiping Li}
\affiliation{%
  \institution{Renmin University of China}
}
\email{licuiping@ruc.edu.cn}




\begin{abstract}
Large Language Models (LLMs) have been increasingly adopted in Text-to-SQL systems, yet SQL errors remain a major obstacle in real-world Text-to-SQL inference pipelines. Existing SQL correction approaches either rely on large-scale, high-quality training data with substantial overhead, or adopt single-path agentic workflows that are brittle to early mistakes and prone to error propagation.

To develop a practical SQL correctness system for industrial scenarios, we present a training-free framework that formulates SQL correction as a plan-guided, tree-structured debugging process. By maintaining multiple correction strategies and enabling backtracking, the framework mitigates error accumulation during iterative refinement. We further integrate execution-based verification and clause-level diagnostic tools to support strategy pruning and precise error localization.

We evaluate the system on the BIRD-Critic benchmark and observe consistent accuracy gains over strong LLM backbones and representative agent-based baselines, achieving a 9.42\% improvement over the previous state-of-the-art method. The framework is also deployed in the Torch Log Service (TLS) of Volcano Engine to support an online Text-to-TLS API. In production, it improves execution accuracy from 36.77\% to 53.61\% on real user queries with a representative strong LLM backbone (GPT-5). These results demonstrate the effectiveness and stability of our approach in real-world deployments.
\end{abstract}

\maketitle

\pagestyle{\vldbpagestyle}
\begingroup\small\noindent\raggedright\textbf{PVLDB Reference Format:}\\
\vldbauthors. \vldbtitle. PVLDB, \vldbvolume(\vldbissue): \vldbpages, \vldbyear.\\
\href{https://doi.org/\vldbdoi}{doi:\vldbdoi}
\endgroup
\begingroup
\renewcommand\thefootnote{}\footnote{\noindent
This work is licensed under the Creative Commons BY-NC-ND 4.0 International License. Visit \url{https://creativecommons.org/licenses/by-nc-nd/4.0/} to view a copy of this license. For any use beyond those covered by this license, obtain permission by emailing \href{mailto:info@vldb.org}{info@vldb.org}. Copyright is held by the owner/author(s). Publication rights licensed to the VLDB Endowment. \\
\raggedright Proceedings of the VLDB Endowment, Vol. \vldbvolume, No. \vldbissue\ %
ISSN 2150-8097. \\
\href{https://doi.org/\vldbdoi}{doi:\vldbdoi} \\
}\addtocounter{footnote}{-1}\endgroup

\ifdefempty{\vldbavailabilityurl}{}{
\vspace{.3cm}
\begingroup\small\noindent\raggedright\textbf{PVLDB Artifact Availability:}\\
The source code, data, and/or other artifacts have been made available at \url{\vldbavailabilityurl}.
\endgroup
}

\begin{figure*}[t]
    \centering
    \includegraphics[width=0.95\textwidth]{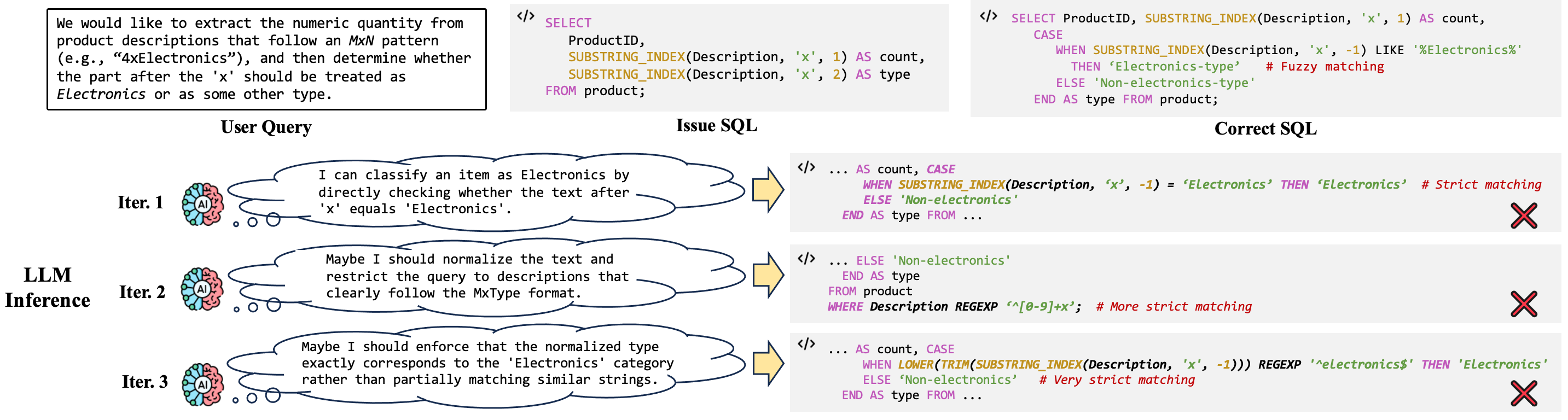}
	\caption{\label{fig:intro_case} A failure mode of linear SQL correction, illustrating how early implicit assumptions and error accumulation lead to semantic drift, causing the final query to deviate from the user's intent.}
\end{figure*}

\section{Introduction}
\label{sec:intro}
Relational databases underpin modern data-driven applications, with SQL serving as the primary interface for querying structured data. However, writing correct SQL remains difficult even for experienced users. Beyond syntactic proficiency, it requires precise reasoning over complex schemas, join relationships, aggregation semantics, and predicate logic. In large real-world databases, minor mistakes—such as incorrect join conditions or misplaced filters—can silently produce logically incorrect results. 

Large Language Models (LLMs) have substantially advanced Text-to-SQL systems, significantly reducing the burden of manual query writing. However, generating correct SQL in a single pass remains challenging. Even strong base and fine-tuned models frequently produce syntax errors, incorrect join paths, aggregation mismatches, and other semantic inconsistencies.
For example, on the Text-to-SQL benchmark Spider2.0~\cite{spider2}, a strong base model such as Claude-4~\cite{claude4} achieves only 25.78\% accuracy. In industrial products, such as Torch Log Service (TLS) on Volcano Engine, where databases are highly heterogeneous and customized SQL dialects are common, such errors occur even more frequently.

These challenges highlight the necessity of SQL correction beyond one-pass generation. Recent agentic approaches attempt to address this issue by incorporating multi-round interaction and execution-guided correction into the inference process~\citeN{dac, alphasql}. However, the correction module in these approaches typically rely on execution-guided refinement or agent-style self-debugging, but often follow single-path reasoning and are vulnerable to early errors or limited generalization. As a result, achieving reliable and scalable SQL issue resolution remains an open problem.



\vpara{Real-World Deployment: Applications \& Constraints.}
With the growing adoption of large language models (LLMs) for Text-to-SQL~\citeN{codes, alphasql, survey}, SQL correctness has become critical module in real-world pipelines of Text-to-SQL applications based on LLMs. In industrial LLM-based Text-to-SQL systems in ByteDance, automated SQL correction is critical in two stages. During training, large-scale datasets are frequently synthesized or augmented using LLM-generated SQL, where uncorrected errors directly degrade data quality. During inference, correcting imperfect SQL outputs before returning to users substantially improves the reliability and usability of online Text-to-SQL APIs. Therefore, effective SQL correction is essential for both training data quality and end-to-end system performance.

Additionally, although training on large-scale in-domain data can significantly enhance LLM performance in specific domains~\citeN{codes, omnisql}, such approaches face challenges in real-world SQL correctness systems. 
First, collecting high-quality supervision for SQL correctness is costly and time-consuming, as it requires realistic user intents, diverse database schemas, and verified ground-truth SQL queries. Second, real-world deployments typically involve multiple SQL dialects, whereas models fine-tuned on limited dialects often exhibit degraded generalization when transferred to unseen ones. For example, we observe that the Xiyan model~\cite{xiyan}, trained on the maintain dialect such as MySQL, PostgreSQL and so on, achieves 13.40\% accuracy on our real-world Torch Log Service (TLS) benchmark, while the corresponding base model without additional fine-tuning (GPT-5~\cite{gpt5}) attains 36.77\%. These limitations substantially hinder the scalability and practicality of training-based solutions, motivating the need for a training-free SQL correctness framework that can robustly adapt to diverse SQL environments.

\vpara{Limitations of Existing Methods.}
Recent training-free SQL correction methods commonly adopt agent-based workflows, where LLMs iteratively refine SQL queries based on execution feedback~\citeN{cscsql, dac, sqlfixagent, qrhint, sqlens, wang2024tool}. In this line of work, the correction process typically follows a linear refinement paradigm, in which the agent revises a single SQL candidate step by step, guided by runtime errors or execution results.
However, this paradigm inherently constrains the agent to a single correction trajectory. As a consequence, the overall process becomes fragile, as it heavily depends on early decisions, and unreliable, since errors introduced at early stages tend to accumulate and persist throughout subsequent iterations.
We illustrate the intrinsic limitations of linear refinement through the following example.

\noindent{\textsc{Example.}}
Figure~\ref{fig:intro_case} presents a representative failure mode of linear, single-path SQL correction. The task is to extract a numeric quantity and classify product types from textual descriptions following an ``$M \times N$'' pattern, where $M$ denotes a number and $N$ denotes a product type string. Notably, the user intent only requires identifying products whose descriptions contain the keyword ``Electronics'', rather than enforcing an exact or canonicalized match (see the correct SQL in the figure). This illustrates the limitations of linear refinement:

\textit{(1) Fragility caused by early implicit assumptions.}
In the first correction step, the model implicitly assumes that identifying ``Electronics'' requires an exact string match. Once this assumption is adopted, all subsequent revisions operate within the same restricted hypothesis space, focusing on refining string normalization and matching strictness without reconsidering whether exact equality is semantically appropriate. As a result, alternative plausible interpretations—such as substring containment—are never explored. This example shows that linear refinement is highly sensitive to early assumptions: a single misinterpretation can dominate the entire correction process. Thus, the linear refinement process could be fragile due to the sensitivity of such implicit assumptions.

\textit{(2) Unreliability due to error accumulation and irreversibility.}
As illustrated by this example, the model progressively introduces increasingly restrictive constraints, such as string normalization (lowercasing and trimming) and regex-based exact matching. These modifications are cumulatively applied to the SQL query, causing the revised query to deviate substantially from its original structure and intent.
In SQL tasks, seemingly local changes—such as adding a filtering condition or tightening a predicate—often have global semantic implications, since they directly affect the result set produced by query execution. When such clause-level modifications are repeatedly layered on top of the SQL query being modified, the compounded effect can cause the final query to drift significantly away from the user’s true intent and initial SQL query.
As a result, linear, iterative correction processes are highly susceptible to error accumulation and semantic drift, which fundamentally undermines their reliability for SQL debugging.


\vpara{Our Proposal.}
The limitations discussed above highlight a fundamental requirement for SQL correction: once semantic deviations are introduced, the system must be able to revert to earlier steps, rather than continuing refinement along a single irreversible path.
To meet this requirement, we adopt tree-based reasoning as an alternative to linear SQL refinement. Specifically, we formulate SQL correction as a tree-structured debugging process, where different critic strategies and semantic hypotheses are explicitly modeled as separate branches, and earlier modification nodes can be revisited when errors are detected.
This formulation mitigates both the fragility caused by early assumptions and the unreliability arising from error accumulation in linear refinement.
Such a tree-structured process relies on two key mechanisms.
\textbf{(1) Branching} preserves alternative correction strategies when multiple revision plans are plausible, for example under ambiguous user intent or underspecified natural language queries.
\textbf{(2) Backtracking} enables the system to discard erroneous paths once inconsistencies are identified, instead of propagating their effects forward through the entire correction trajectory.

Effectively realizing these mechanisms places higher demands on the model’s reasoning and verification capabilities. In particular, tree-based SQL debugging requires the LLM to support the following capabilities.
\textbf{Capability (i) Proposing multiple candidate correction paths}, which is essential for branching, as the quality and diversity of proposed alternatives directly determine whether the correct solution space is preserved. This includes recognizing potential ambiguities in user intent and enumerating reasonable interpretations instead of committing to a single implicit assumption.
\textbf{Capability (ii) Accurately assessing SQL correctness} is critical for backtracking, since identifying semantic mismatches and determining when a branch should be abandoned form the core of recovering from erroneous paths. Notably, syntactically valid SQL may still fail to satisfy user intent, making execution success alone insufficient for reliable validation.
\textbf{Capability (iii) Rapidly recovering from syntactic errors while preserving semantic consistency} ensures that tree exploration can proceed without being stalled by non-executable intermediate queries. Efficient syntactic repair allows the framework to resume semantic verification and branching decisions promptly, thereby maintaining the effectiveness of both branching and backtracking.

To enable tree-structured SQL debugging, we propose Agentic and Critic-Oriented Tree-Structured SQL Correctness (\model), a framework that organizes SQL correction as a plan-guided execution process and provides a set of specialized tools to support efficient and robust debugging. Given a user query and a faulty SQL, the LLM first generates a revision plan that decomposes the correction process into a sequence of tool invocations. This plan is structured as a tree, where each node corresponds to a specific diagnostic or correction operation, and branching points are introduced whenever multiple plausible interpretations or revision strategies arise. Serving as a high-level blueprint, the tree-structured plan enables systematic exploration of alternative hypotheses and controlled rollback when semantic deviations are detected.

The tree-based formulation naturally supports the LLM in proposing multiple candidate correction paths. To further enhance this capability (Capability (i)), we introduce a \textit{Detect Ambiguities} tool that explicitly analyzes underspecified user intents and enumerates multiple plausible semantic hypotheses as branches of the tree. In addition, accurately assessing SQL correctness (Capability (ii)) requires grounding validation in concrete database instances rather than relying solely on LLM reasoning. In SQL tasks, the correctness of joins, filters, and aggregation strategies often depend on data distributions or formats. To address this, we equip the framework with database-interaction tools, including \textit{Run SQL} for execution-based validation and schema-aware inspection operators such as \textit{Inspect Columns}.
Finally, to enable rapid recovery from syntactic failures (Capability (iii)), we introduce a \textit{Split and Fix Syntax Error SQL} tool that decomposes erroneous queries into clause-level components and precisely localizes syntax errors. By repairing syntax errors at the clause level instead of regenerating the entire query, the framework restores executability with minimal disruption to previously validated semantics.

\vpara{Evaluation \& Application.}
We evaluate our framework on the BIRD-Critic benchmark, a comprehensive dataset covering multiple SQL dialects and diverse levels of query errors. Experimental results demonstrate that our method achieves state-of-the-art performance, surpassing the previous best system by 9.42\%.

Beyond public benchmarks, we further validate our approach on a real-world Text-to-TLS dataset collected from an industrial log analytics system in Volcano Engine. In production deployment within the Text2TLS pipeline, using a representative strong LLM backbone (GPT-5), our framework improves execution accuracy by 16.84 percentage points on real user queries under practical latency and cost constraints. Similar improvement trends are observed across other LLM backbones, indicating that the effectiveness of our framework generalizes beyond a specific model. These results demonstrate the robustness and deployability of our approach in real-world environments.

\vpara{Contributions.}
Our main contributions are summarized as follows:
\begin{itemize}[leftmargin=1em]
\setlength\itemsep{0em}
\item We introduce a revision-plan–based debugging paradigm that structures the SQL correction process into a hierarchical, decision-guided plan. This paradigm better aligns with human debugging behavior and significantly enhances interpretability and efficiency.

\item We design an enriched operator space comprising specialized tool interfaces for ambiguity detection, sub-query decomposition, and schema inspection. These operators effectively assist LLMs in addressing both syntactic and semantic challenges frequently encountered in real-world SQL debugging.

\item We conduct extensive experiments on the BIRD-Critic benchmark, demonstrating that our approach achieves state-of-the-art performance, outperforming the previous best fine-tuned model by 9.42\%, thus validating the effectiveness of our framework. Moreover, when deployed in an industrial Text-to-TLS system with a representative strong LLM backbone (GPT-5), our framework improves execution accuracy from 36.77\% to 53.61\%, demonstrating the robustness and general applicability of our method in real-world settings.

\end{itemize}

\begin{figure*}[t]
    \centering
    \includegraphics[width=0.8\textwidth]{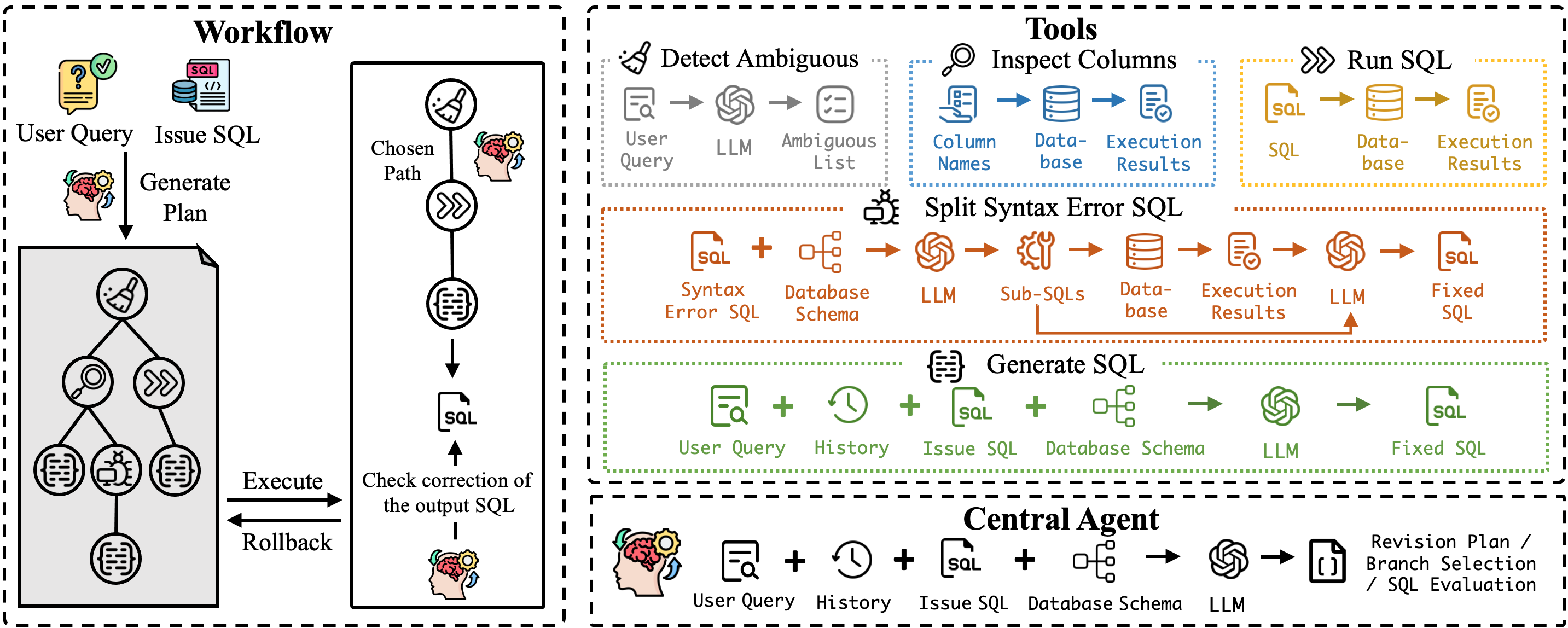}
	\caption{\label{fig:overview} Overview of our framework.}
\end{figure*}

\section{Related Work}
\subsection{Text-to-SQL System}
Recent research has explored agentic reasoning and test-time scaling for Text-to-SQL systems. The recent survey~\cite{survey} finds that both reasoning-specialized and general-purpose LLMs consistently benefit from test-time scaling strategies such as divide-and-conquer~\cite{divide}, few-shot prompting, and ReAct-style reasoning~\cite{react}, leading to robust performance improvements across different model families.

Several works, including LearNAT~\cite{learnat}, MAG-SQL~\cite{magsql}, MAC-SQL~\cite{macsql}, and CoE-SQL~\cite{coesql}, adopt a compositional reasoning paradigm in which each Common Table Expression (CTE) or subquery corresponds to an intermediate reasoning step. These approaches decompose SQL generation into hierarchical subgoals, improving interpretability and controllability. Although these approaches incorporate SQL refinement or revision modules, the refinement process itself typically follows a one-pass or linear trajectory.

Alpha-SQL~\cite{alphasql} also introduces a tree-structured framework, modeling schema linking, SQL generation, SQL revision and other sub-tasks as nodes in a hierarchical reasoning process. However, its tree structure primarily organizes generation stages rather than exploring alternative correction branches. In particular, its revision component refines SQL along a single trajectory.

\subsection{SQL Revision}
SQL debugging approaches can be broadly categorized into static analysis and AI-based refinement.
Traditional static tools—such as UPM~\cite{upm}, Visual Expert~\cite{visualexpert}, SQLFluff~\cite{sqlfluff}, and SQLCheck~\cite{sqlcheck}—apply rule-based analysis on Abstract Syntax Trees (ASTs) to identify anti-patterns, type inconsistencies, and inefficient query structures. While these systems provide reliable diagnostics, they lack semantic awareness of user intent and cannot adaptively repair queries.

In contrast, AI-based refinement methods leverage LLMs to iteratively inspect, execute, and modify SQL queries~\citeN{ambisql, maplerepair, sqlcritic}.
SQLFixAgent~\cite{sqlfixagent} employs a “rubber-duck debugging” paradigm, generating perturbed query variants to explore diverse semantic hypotheses.
CSC-SQL~\cite{cscsql} improves candidate selection through collaborative training between a SQL generator and a merger guided by execution feedback.
To enhance correction precision, recent studies introduce decomposition-based refinement:

\ipara{Tool-based decomposition.} Tool-Assisted Agent~\cite{wang2024tool} defines modular functions for modifying SQL clauses (e.g., JOIN, WHERE, GROUP) and provides validity feedback for each action; CoE-SQL~\cite{coesql} models AST editing as sequential reasoning steps; Qr-Hint~\cite{qrhint} introduces actionable hints for error localization.

\ipara{Clause-based decomposition.} SQLCritic~\cite{sqlcritic} contrasts incorrect–correct clause pairs to train a critic model for clause-level quality assessment.

\ipara{Subtask-based decomposition.} DAC~\cite{dac} decomposes the overall refinement process into entity linking, skeleton parsing, and SQL validation, aligning task-specific feedback with the final correction.

Despite these advances, most existing frameworks rely on linear refinement loops, which suffer from error propagation and limited exploration capability when debugging complex queries.

\subsection{Code Revision}
The idea of iterative self-improvement has been widely studied in the code refinement domain.
Systems such as PAG~\cite{pag}, CYCLE~\cite{cycle}, LeDex~\cite{ledex}, QualityFlow~\cite{qualityflow}, AdaCoder~\cite{adacoder}, RefineCoder~\cite{refinecoder}, and CodeCoR~\cite{codecor} employ cyclic refinement pipelines in which the model repeatedly executes, inspects, and revises generated code. These frameworks demonstrate the effectiveness of execution-guided feedback and self-reflective reasoning, but their repair trajectories remain single-threaded and prone to cumulative errors.
To enhance controllability, several works incorporate external tool integration: RepairAgent~\cite{repairagent}, CodeAgent~\cite{codeagent}, and LANTERN~\cite{lantern} enable the LLM to decide when to invoke specialized repair or verification tools. Notably, MGDebugger~\cite{mgdebugger} constructs a hierarchical debugging tree (syntax → function → algorithm) that progressively refines the program through structured iterations, while Divide \& Conquer Revision~\cite{divide} separates error localization from repair.

However, when directly applied to SQL debugging, these frameworks face additional challenges: SQL’s symbolic compositionality and semantic dependencies (e.g., among SELECT, JOIN, and WHERE clauses) make single-path refinement unstable. This motivates our approach, which introduces an explicit revision plan to organize the debugging process into a tree-structured reasoning space, enabling multi-branch exploration, dynamic tool feedback integration, and interpretable refinement trajectories.

\nocite{*}
›\section{System Overview}
SQL correction in practical environments is challenging due to the tightly coupled structure of SQL queries, the complexity and variability of real-world database schemas, and the inherent ambiguity in user intent etc. 
Our framework addresses these challenges by structuring SQL correction as a revision-plan–guided process and equipping the model with specialized operators for ambiguity detection, sub-query decomposition, and schema inspection etc. This design enables systematic debugging, improves correction reliability across diverse scenarios, and reduces the human effort required in both development workflows and Text-to-SQL production systems.

\subsection{Workflow}
The system operates through a structured, plan-driven workflow which begins with the LLM first producing an initial revision plan according to the given user query and the faulty SQL. The plan can be described as a tree, each node of the tree specifies a tool invocation and potential branching paths depending on the tool’s return signals. This plan acts as a high-level blueprint that decomposes the debugging process into interpretable and actionable steps.

After producing an initial plan, the system executes it by calling predefined tools and allowing the \brain to decide the next branch based on returned tool results. The plan can expand dynamically whenever a tool introduces new decision points. When a tool returns multiple candidate branches—for example, ambiguity hypotheses from the detect-ambiguities tool—the controller selects the most credible one and continues execution along that path.
Once a complete SQL candidate is generated, the \brain evaluates its correctness based on the histories and execution feedback. If it is incorrect, the system performs a bottom-up rollback to explore alternative paths. The process ends when the controller finds a correct SQL.

\subsection{Components}
Our framework consists of two main components—\textit{\brain} and \textit{Tools}, which including five main useful tools \textit{Detect Ambiguous}, \textit{Inspect Columns}, \textit{Run SQL}, \textit{Split and Fix Syntax Error SQL} and \textit{Generate SQL}. These components collaboratively perform the SQL refinement task as follows:

\vpara{\brain.}
\brain serves as the ``brain'' of our framework. It is responsible for generating the initial revision plan, selecting the appropriate branch whenever the plan diverges, and evaluating the correctness of the produced SQL. The agent takes as input the user query, database schema, issue SQL, and relevant execution history, and outputs decisions that drive each stage of the correction process. Through these responsibilities, the \brain coordinates the entire system, ensuring that SQL refinement progresses coherently and converges toward a correct or high-confidence solution.

\vpara{Tools.}
Tools define the operator space of our framework, providing the actionable steps through which the \brain refines SQL. The tools include: LLM-based tools such as for users' intent interpretation, execution tools for running SQLs and returning feedback, and hybrid tools that combine reasoning with execution. Together, they provide the exploration space that enables iterative and reliable SQL correction.

\ipara{Detect Ambiguous(LLM-based tool).}
User queries often contain ambiguous expressions. For example, in the StackOverflow database, the term “active users” could refer to users who post, users who answer, or pre-mine users. This tool takes the raw user query as input and uses an LLM to identify potentially ambiguous phrases, along with plausible interpretive assumptions for each.


\ipara{Inspect Columns(execution tool).}
Column values can guide SQL correction by revealing actual data patterns, which aids in crafting precise matching predicates. This tool provides an interface to retrieve sample values from specified database columns.

\ipara{Run SQL(execution tool).}
In practice, human data analysts often execute sub-SQLs to inspect intermediate query results or run simplified queries to validate SQL syntax. This tool provides flexibility by allowing the execution of any given SQL query and returning the corresponding execution result, supporting the debugging process by facilitating query validation.

\ipara{Split and Fix Syntax Error SQL(hybrid tool).}
To improve the efficiency of fixing syntax errors, this tool leverages the LLM to decompose error SQL into smaller, independent subqueries. These subqueries can then be executed in parallel, reducing the time spent on error identification and correction. The execution results, along with the subqueries, are subsequently fed back into the LLM for further refinement, ultimately returning a revised SQL query free from syntax errors.

\ipara{Generate SQL(hybrid tool).}
This tool appears as a leaf node in the tree-structured revision plan to output the final SQL. After reasoning by the \brain and the execution of intermediate tools, the accumulated history, along with the user query, original SQL, and database schema are provided to the LLM. The LLM generates the corrected SQL query and returns both the revised query and its execution result, completing the SQL correction process.

\begin{figure}[t]
    \centering
    \includegraphics[width=0.488\textwidth]{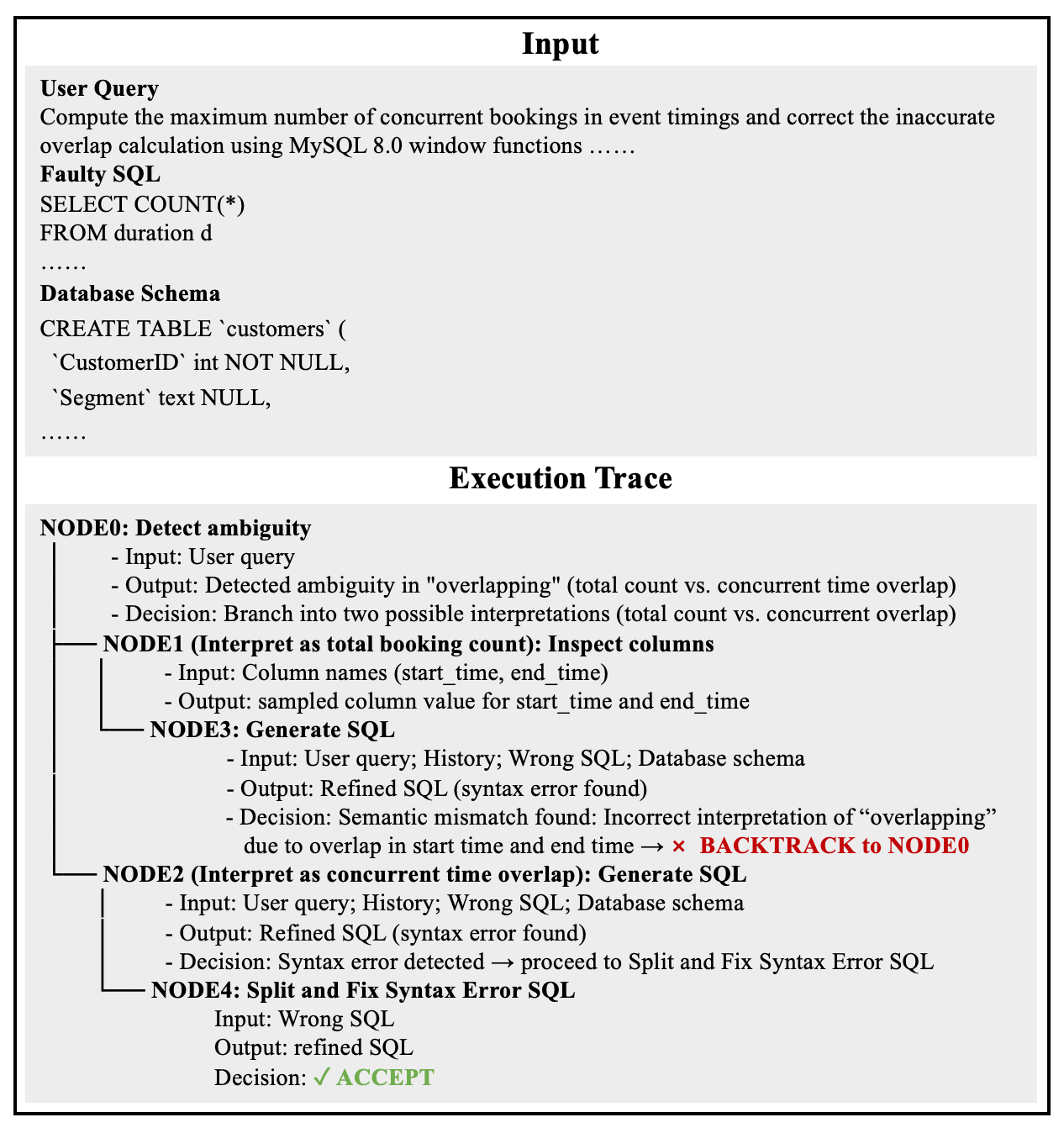}
	\caption{\label{fig:workflow_case} An example of the tree-based SQL correction pro- cess. Starting from ambiguity detection, the system branches into alternative semantic interpretations of ``overlapping''. The incorrect interpretation (total booking count) is pruned after semantic validation based on evidence obtained from the inspect\_columns tool, triggering backtracking. The cor- rect branch (concurrent time overlap) proceeds through SQL generation and syntax repair, ultimately producing a validated query. Each node corresponds to a tool invocation, and branching, backtracking, and acceptance decisions are made by Central Agent based on execution feedback.}
\end{figure}

\section{\model}
In this section, we introduce our plan-guided revision framework. We first detail how the system generates, expands, and executes the tree-structured revision plan. We then introduce the \brain, including its input prompt design and the outputs that orchestrate the overall correction process.

\subsection{Plan Generation and Execution}
The non-linear debugging workflow in our framework is represented as a tree in which each node corresponds to a tool invocation and each edge denotes a potential follow-up action. The overall procedure is summarized in Algorithm \ref{alg:tree_debug}.

\subsubsection{Plan Generation}
Given the user issue, SQL dialect, database schema, problematic SQL query, and its execution result, the system first constructs an initial revision plan (line~1). This plan specifies the root diagnostic operation and enumerates major branches, each representing a distinct interpretation of the error or a plausible repair direction. The debugging process then starts from the root node (line~2).

\begin{algorithm}[t]
\caption{Tree-Structured SQL Debugging}
\label{alg:tree_debug}
\KwIn{User issue $U$, initial SQL $S_{0}$, database schema $D$}
\KwOut{Corrected SQL $S^{*}$}

$T \leftarrow \textsc{GenerateInitialPlan}(U, S_{0}, D)$ \;
$n \leftarrow T.\text{root}$ \;

\While{\texttt{True}}{

    $result \leftarrow \textsc{Execute}(n)$ \;

    \If{$n.\text{tool} = \textsc{detect\_ambiguities}$}{
        $n \leftarrow \textsc{ExpandTree}(T, result)$ \;
    }

    \If{$|n.\text{children}| > 1$}{
        $n \leftarrow \textsc{SelectNextNode}(result)$ \;
        \Continue \;
    }

    \If{$n.\text{tool} = \textsc{generate\_sql}$}{
        \If{\textsc{ValidSQL}(result)}{
            \Return $result$ \;
        }
        \Else{
            $n \leftarrow \textsc{Rollback}(T, n)$ \;
            \Continue \;
        }
    }

    \If{$|n.\text{children}| = 1$}{
        $n \leftarrow n.\text{children}[0]$ \;
    }
}
\end{algorithm}

\subsubsection{Dynamic plan expansion}
When the current tool detects ambiguity in the user’s natural-language query, it expands the current node by generating additional child branches that represent alternative hypotheses (lines~5–7). Each child node corresponds to one such hypothesis, and \brain then selects the most plausible branch for continued reasoning, following the same decision mechanism described in the next subsection.

\subsubsection{Choose Next Node to Execute}
During execution, the system iteratively selects the next node to traverse based on the outcomes of previously executed steps. When the current node contains multiple child nodes, \brain chooses the most promising one according to the intermediate results produced at that node (lines~8–11). This iterative cycle of traversal and selective expansion enables the system to adapt its reasoning path as additional evidence is gathered.

\subsubsection{Terminal of the Execution}
Execution terminates when a leaf node yields a SQL candidate that satisfies the \brain's validation checks. Once the \texttt{generate\_sql} tool is invoked, a candidate SQL is produced and \brain evaluates its correctness (line~13). If the candidate passes validation, the debugging process terminates. Otherwise, \brain may backtrack to any node in the plan and continue the revision process.

\subsection{\brain}
As the ``brain'' of our framework, \brain serves as the central agent responsible for generating and orchestrating the execution of the revision plan. Figure~\ref{fig:workflow_case} illustrates an example of the overall workflow. Specifically, \brain generates the initial revision plan, selects branches at $NODE\ 0$, and evaluates the SQL produced at $NODE\ 2$, $NODE\ 3$, and $NODE\ 4$ to determine whether to continue, backtrack, or accept the result.
To support these responsibilities, we design the prompt for \brain to include the following components to provide comprehensive information:

\begin{itemize}[leftmargin=1em]
\setlength\itemsep{0em}
\item \textbf{Task Overview \& Instruction.}
This section defines the agent’s core objectives and responsibilities, outlining high-level principles such as when to construct an initial plan, how to select subsequent nodes, and how to determine termination conditions.

\item \textbf{Available Tools.}
We provide detailed specifications for all tools accessible during debugging, including their purposes, recommended usage scenarios, and input–output formats. This enables \brain to reason over a well-defined operator space when revising queries.

\item \textbf{Output Format.}
The \brain is responsible for three core tasks: (i) constructing the tree-structure revision plan, (ii) selecting the next node to execute when branching occurs, and (iii) determining whether the SQL candidate is correct upon reaching a leaf node. To unify these control actions, we formalize the entire process as operations over a JSON object. The initial plan generation corresponds to producing this JSON structure, whereas intermediate decisions correspond to updating it. Each plan node—corresponding to a single tool invocation—is represented as a dictionary containing a unique node identifier, a short textual description, the tool name, the tool input, and a list of child nodes representing possible follow-up actions. To enable dynamic control flow, we introduce a global \texttt{cursor} field that marks the node currently being executed; \brain perform task (ii) by updating this field to point to the next selected node.

\item \textbf{Input Information.}
\brain receives comprehensive contextual information to support accurate, stateful reasoning. This includes the SQL dialect, database schema, user-issued problem description, the problematic SQL, and its execution results. We also provide the complete interaction history—the current plan (in JSON form) and the outputs of previously executed nodes. Executed nodes are annotated with their corresponding execution results, and the \texttt{cursor} is updated accordingly to reflect the most recent execution event, ensuring coherent and context-aware reasoning throughout the revision process.
\end{itemize}

\subsection{Details of Tools}
The tool set defines the actionable space through which \brain explores and refines the debugging trajectory. 
To address the limitations in resolving syntax and semantic error SQLs of existing methods mentioned in Section~\ref{sec:intro}, we construct a tool space that accelerate syntax error SQL fixing and detect semantic mismatches. Specifically, tools such as \textsc{Detect Ambiguities}, \textsc{Run SQL}, and \textsc{Inspect Column Format} allow the agent to systematically probe the mismatch between user intent and SQL behavior, providing crucial signals for resolving semantic inconsistencies. Meanwhile, \textsc{Split Syntax Error SQL} decomposes faulty queries into clause-level substructures, enabling parallel validation and rapid localization of multiple syntax issues within a single iteration.

This section presents the tools used in our framework. We first describe tools for detecting semantic ambiguity, followed by tools that execute or inspect SQL behavior. We then introduce tools designed for structural decomposition of syntax error SQL, and finally, the \textsc{Generate SQL} tool that generate corrected SQL candidates.

\subsection{Detect Ambiguities}
The \textsc{Detect Ambiguities} tool is an LLM-based semantic diagnostic module that identifies unclear or underspecified phrases in the user’s natural-language request. Specifically, the prompt includes task description, an example and input information including database schema, user issue and wrong SQL. Thanks to that, the tool highlights ambiguous intent terms and provides 2–3 plausible schema-grounded interpretations for each.

These interpretations become semantic branches in the debugging tree (e.g., NODE 0 in Figure~\ref{fig:workflow_case}), enabling the agent to explore and verify different meanings during later execution. By making implicit ambiguities explicit, this tool provides the foundation for resolving semantic errors that cannot be detected by SQL execution alone.

\subsection{Execution Tools}
The execution tools allow \brain to directly interact with the database and obtain reliable behavioral feedback during debugging. They provide the signals needed to validate assumptions, evaluate intermediate hypotheses, and other that cannot be inferred from static SQL analysis.

\textsc{Run SQL}.
This tool executes any SQL statement proposed by \brain and returns the resulting tuples in list form. It enables the agent to test semantic assumptions, validate sub-queries, and check whether a candidate SQL behaves consistently with the intended logic and so on.

\textsc{Inspect Column Format}.
To support lightweight schema probing, this tool takes a table name and column name provided by \brain and fills them into a fixed query template (e.g., SELECT `column` FROM `table`). It then returns the sampled values in that column. This allows the agent to get data format without manually constructing full queries.

\subsection{Split and Fix Syntax Error SQL}
The Split and Fix Syntax Error SQL tool is invoked when a syntactic error is detected in the generated SQL (e.g., NODE 4 in Figure~\ref{fig:workflow_case}).
It decomposes a syntactically invalid SQL query into a set of minimal, independently executable sub-SQL units. We first prompt the LLM to rewrite the faulty query into primitive relational forms—such as table projection, table filtering, table join, or table aggregation—each consisting of exactly one flat SELECT statement without nested structures.

This decomposition preserves the original semantics while exposing an explicit dependency structure among the generated sub-SQLs. Since this tool focuses solely on syntax validation, we create empty temporary tables to enable parallel execution. For instance, if SQL $B$ depends on SQL $A$, we create a temporary table named $A$ whose schema matches the output columns of SQL $A$, allowing both statements to be executed independently for syntax checking.

Importantly, the generated sub-SQL statements together with their execution results (e.g., syntax errors or validation signals) are fed back to the LLM as structured diagnostic feedback. Based on this fine-grained clause-level information, the LLM synthesizes a revised SQL statement that resolves the detected syntax errors while preserving the intended semantics.
By isolating syntax violations at the clause level and exposing intermediate relational structures, this tool provides precise diagnostic signals that guide the subsequent correction process toward generating a syntactically valid final query.

\subsection{Generate SQL}
The Generate SQL tool is invoked to produce a candidate corrected SQL query and typically serves as a leaf node in the correction tree (e.g., NODE 2 and NODE 4 in Figure~\ref{fig:workflow_case}). 
After generation, the SQL is automatically executed, and the SQL is evaluated by \brain to determine whether to accept the query or trigger further refinement (e.g., backtracking or syntax repair).
To generate the repaired SQL, the LLM receives the full contextual state—including the user’s intent description, the database schema, the problematic SQL, execution feedback, and diagnostic signals accumulated during earlier tool invocations from root to current node—and produces a revised SQL query that is both syntactically valid and semantically aligned with the user’s goal.

In this tool, the LLM is instructed to minimally edit the original query, avoid speculative assumptions, and ground its modifications strictly in the provided evidence. This design ensures that the final output is a faithful, precise correction rather than an overly rewritten or hallucinated alternative. Once generated, the SQL candidate is executed to confirm correctness, completing the debugging workflow.

\section{Experimental Evaluation}
We evaluate the proposed system through a comprehensive set of experiments designed to assess its effectiveness, robustness, and practicality. We first report benchmark results to quantify overall performance gains, followed by ablation studies to analyze the contribution of key components. We then present results on a real-world Text-to-SQL (TLS) benchmark to demonstrate performance under practical system constraints. In addition, we provide representative case studies to illustrate the system’s correction behavior in complex scenarios, analyze runtime cost and latency characteristics, and finally discuss its application in a real-world production deployment.
\begin{table*}[t]
    \centering
    \caption{Main results on the BIRD-CRITIC benchmark.}
    \label{tab:bird_results}
    \footnotesize
    \setlength{\tabcolsep}{5pt}
    \renewcommand{\arraystretch}{0.95}
    \begin{tabularx}{\textwidth}{lY|YYYYY}
        \toprule
        \multicolumn{2}{l}{\textbf{Methods}} 
        & \multicolumn{4}{c}{\textbf{Bird-critic Open}} 
        & \textbf{Bird-critic PG} \\
        \cmidrule(lr){3-6}
        \multicolumn{2}{l}{} 
        & PostgreSQL & MySQL & SQLServer & Oracle &  \\
        \midrule
        \multirow{4}{*}{Raw model}
        & Deepseek-R1~\cite{deepseekr1} & 36.96\% & 38.78\% & 32.65\% & 19.39\% & 39.62\% \\
        & GPT-o3~\cite{gpt_o3}      & 41.30\% & 26.53\% & 32.65\% & 18.37\% & 38.87\% \\
        & GPT-5~\cite{gpt5}       & 39.13\% & 28.57\% & 29.59\% & 23.71\% & 35.85\% \\
        & grok-4~\cite{grok4}      & 39.49\% & 35.71\% & 33.67\% & 15.31\% & 39.06\% \\
        \midrule
        \multirow{4}{*}{Agents} 
        & SQLFixAgent~\cite{sqlfixagent} & 21.38\% & 22.79\% & 33.12\% & 17.68\% & 40.56\% \\ 
        & DAC~\cite{dac}         & 19.00\% & 16.00\% & 18.00\% & 3.00\% & 31.32\% \\ 
        & RepairAgent~\cite{repairagent} & 34.94\% & 33.67\% & 34.20\% & 28.12\% & 35.27\% \\ 
        & MapleRepair~\cite{maplerepair} & 44.20\% & 37.76\% & 39.79\% & 22.45\% & 37.74\% \\ 
        \midrule
        \multicolumn{2}{Y|}{Xiyan Model~\cite{xiyan}} 
        & \underline{49.28\%} & \underline{45.92\%} & \underline{43.88\%} & \underline{28.57\%} & \underline{44.53\%} \\ 
        \midrule
        \multicolumn{2}{Y|}{\textbf{\model}}
        & \textbf{58.70\%} & \textbf{64.29\%} & \textbf{53.06\%} & \textbf{33.67\%} & \textbf{45.85\%} \\
        \bottomrule
    \end{tabularx}
\end{table*}

\subsection{Evaluation on Benchmarks}
In this section, we evaluate \model on the public BIRD-CRITIC benchmark~\cite{swesql} and compare it with representative baselines to validate its effectiveness in SQL debugging.

\subsubsection{Evaluation Settings}
We describe the benchmarks, baselines, and evaluation metrics used in this experiments below.

\ipara{Benchmark - BIRD-CRITIC.}
We evaluate \model on \textbf{BIRD-CRITIC}, a benchmark for SQL issue debugging built from real-world user-reported SQL errors~\cite{swesql}.
It contains a PostgreSQL subset (BIRD-CRITIC-PG) and a multi-dialect subset spanning four database dialects.
The benchmark is highly challenging and targets systematic SQL debugging rather than SQL generation.
We report results on both the PostgreSQL and the open multi-dialect subsets.

\ipara{Baselines.}
We compare our approach with representative baselines covering different SQL tasks.

\begin{itemize}[leftmargin=1em]
\setlength\itemsep{0em}
\item \textbf{Raw Models.}
We evaluate several strong general-purpose LLMs as raw baselines, including GPT-o3~\cite{gpt_o3}, GPT-5~\cite{gpt5}, and DeepSeek-R1~\cite{deepseekr1}.
These models are prompted to directly perform SQL correction without any additional training or external tools.
For all raw models, we use official inference APIs and adopt consistent decoding configurations, with temperature set to 0.1, top-$p$ set to 0.95, and a maximum input length of 8K tokens.

\item \textbf{SQL Correction Agents.}
We include SQLFixAgent~\cite{sqlfixagent}, a trained agent-based method that performs semantic mismatch diagnosis and candidate generation using a rubber-duck debugging strategy.
We also evaluate two training-free SQL correction frameworks, DAC~\cite{dac} and MapleRepair~\cite{maplerepair}, which diagnose and repair SQL errors through sub-task comparison and rule-based symptom detection, respectively, with optional LLM-based regeneration.
All these methods follow a predominantly linear refinement paradigm, where correction decisions are made sequentially without explicitly preserving alternative revision branches.


\item \textbf{Fine-tuned Text-to-SQL Models.}
We include \emph{XiYan-SQL}, a strong fine-tuned Text-to-SQL model trained with multi-task objectives across diverse SQL formats and database dialects~\cite{xiyan}.
XiYan-SQL has demonstrated state-of-the-art performance on standard Text-to-SQL benchmarks such as BIRD and Spider.
For fair comparison, we adopt the same prompting protocol as used in the Bird-Critic leaderboard for evaluating raw models, and apply it consistently across the fine-tuned baselines.

\end{itemize}

\begin{table*}[t]
    \centering
    \caption{Ablation study of \model on the BIRD-CRITIC benchmark.}
    \label{tab:ablation}
    \footnotesize
    \setlength{\tabcolsep}{5pt}
    \renewcommand{\arraystretch}{0.95}
    \begin{tabularx}{\textwidth}{ll|YYYYY}
        \toprule
        \multicolumn{2}{l}{\textbf{Variants}} 
        & \multicolumn{4}{c}{\textbf{Bird-critic Open}} 
        & \textbf{Bird-critic PG} \\
        \cmidrule(lr){3-6}
        \multicolumn{2}{l}{} 
        & PostgreSQL & MySQL & SQLServer & Oracle &  \\
        \midrule
        \multirow{1}{*}{Full Method}
        & \model (Full)
        & \textbf{58.70\%} & \textbf{64.29\%} & \textbf{53.06\%} & \textbf{33.67\%} & \textbf{45.85\%} \\
        \midrule
        \multirow{1}{*}{Plan Structure}
        & w/ Linear Plan
        & 45.81\% & 36.93\% & 38.88\% & 20.47\% & 36.56\% \\
        \midrule
        \multirow{2}{*}{Tools}
        & w/o Execution Tool
        & 39.84\% & 32.66\% & 29.85\% & 21.48\% & 37.92\% \\
        & w/o Ambiguity Detection
        & 43.89\% & 27.37\% & 31.28\% & 25.67\% & 38.45\% \\
        \bottomrule
    \end{tabularx}
\end{table*}

\ipara{Implementation Details.}
All methods are evaluated under the same experimental settings.
For agent-based approaches, we follow the official implementations and configurations released by the authors.
Unless otherwise specified, large language models are used with deterministic decoding settings to reduce randomness during evaluation.
SQL correctness is determined by execution-based evaluation, where a predicted SQL query is considered correct if it produces the same result as the ground-truth query under the target database.

\subsubsection{Results on Benchmark}
We evaluate \model on the BIRD-CRITIC benchmark, which is specifically designed for SQL debugging and correction across multiple SQL dialects.
Table~\ref{tab:bird_results} summarizes the results on both the BIRD-CRITIC-Open (multi-dialect) and BIRD-CRITIC-PG settings and we have the following findings:

\textbf{Compared with raw LLMs,} \model consistently outperforms strong raw and reasoning-based LLMs on both BIRD-CRITIC-PG and the multi-dialect setting.
We attribute this improvement to the fact that SQL debugging is inherently a structured program repair task, which requires precise error localization and iterative verification rather than one-shot generation.
While powerful foundation models exhibit strong reasoning abilities, they typically perform implicit, end-to-end correction, making them brittle when facing complex or compositional SQL errors.
In contrast, our approach externalizes the debugging process into a sequence of structured steps, allowing the model to explicitly test intermediate hypotheses through execution-based feedback and revise SQL in a controlled manner.
These results suggest that better task structuring and feedback integration, rather than stronger base models alone, are crucial for effective SQL issue debugging.


\textbf{Compared with existing agent-based SQL correction methods,} our approach achieves consistently better performance on BIRD-CRITIC.
Although prior agents are also able to execute SQL and leverage execution feedback, they typically follow a linear refinement process.
In practice, linear correction often commits to an early semantic assumption about the error.
When this assumption is incorrect, subsequent edits tend to accumulate errors or repeatedly apply local fixes without addressing the root cause.
\model avoids this issue by structuring SQL correction as a plan-guided process that allows alternative hypotheses to be explored and revised based on execution feedback.
This design makes our agent more robust to misleading intermediate results and better suited for complex SQL debugging scenarios.


\textbf{Compared with a Text-to-SQL model fine-tuned on public SQLite-based datasets.}
While the fine-tuned model achieves strong performance on standard Text-to-SQL benchmarks, it shows limited effectiveness on BIRD-CRITIC, particularly in the multi-dialect setting.
This suggests that fine-tuning on a single dialect and generation-oriented task does not readily transfer to SQL debugging scenarios involving diverse dialects and error patterns.
In contrast, \model improves SQL correctness through a training-free, execution-guided correction process at inference time, making it less dependent on dialect-specific supervision. This results in more consistent performance across PostgreSQL, MySQL, SQL Server, and Oracle, demonstrating good transferability in practical multi-dialect environments.

Overall, \model consistently outperforms strong base models, representative agent-based approaches, and a fine-tuned Text-to-SQL model on BIRD-CRITIC.
The results indicate that structured, execution-guided correction is more effective than linear refinement or dialect-specific fine-tuning, and generalizes better across diverse SQL dialects.

\begin{table*}[t]
    \centering
    \caption{
    Performance of different critic strategies on the TLS dataset.
    Accuracy (\textbf{Acc.}, \%, $\uparrow$) measures exact-match correctness.
    Latency (\textbf{Lat.}, seconds, $\downarrow$) reports the average end-to-end runtime.
    }
    \label{tab:tls_results}
    \footnotesize
    \setlength{\tabcolsep}{5pt}
    \renewcommand{\arraystretch}{1.05}
    \begin{tabularx}{\textwidth}{l|YY|YY|YY|YY|YY}
        \toprule
        \multirow{2}{*}{\textbf{Base Model}}
        & \multicolumn{10}{c}{\textbf{Critic Strategy}} \\
        \cmidrule(lr){2-11}
        & \multicolumn{2}{c}{\textbf{No Critic}}
        & \multicolumn{2}{c}{\textbf{LLM Critic}}
        & \multicolumn{2}{c}{\textbf{DAC~\cite{dac}}}
        & \multicolumn{2}{c}{\textbf{MapleRepair~\cite{maplerepair}}}
        & \multicolumn{2}{c}{\textbf{\model}} \\
        \cmidrule(lr){2-3}
        \cmidrule(lr){4-5}
        \cmidrule(lr){6-7}
        \cmidrule(lr){8-9}
        \cmidrule(lr){10-11}
        & \textbf{Acc.$\uparrow$} & \textbf{Lat.$\downarrow$}
        & \textbf{Acc.$\uparrow$} & \textbf{Lat.$\downarrow$}
        & \textbf{Acc.$\uparrow$} & \textbf{Lat.$\downarrow$}
        & \textbf{Acc.$\uparrow$} & \textbf{Lat.$\downarrow$}
        & \textbf{Acc.$\uparrow$} & \textbf{Lat.$\downarrow$} \\
        \midrule
        GPT-5~\cite{gpt5}
        & 36.77\% & 6.31 
        & \underline{47.42\%} & 26.52 
        & 37.74\% & 72.51 
        & 46.83\% & 15.51 
        & \textbf{53.61\%} & 218.91 \\
        GPT-o3~\cite{gpt_o3}
        & 40.55\% & 18.17
        & \underline{51.89\%} & 35.40 
        & 40.55\% & 139.25 
        & 50.41\% & 17.49 
        & \textbf{56.36\%} & 277.41 \\
        DeepSeek-R1~\cite{deepseekr1}
        & 22.68\% & 6.04 
        & \underline{22.96\%} & 13.24 
        & 22.68\% & 96.04 
        & 22.68\% & 14.55
        & \textbf{23.85\%} & 38.65 \\
        Xiyan~\cite{xiyan}
        & 13.40\% & 7.01 
        & 13.40\% & 15.67 
        & 13.40\%  & 76.54 
        & 13.40\%  & 18.90
        & \textbf{13.75\%}  & 42.56 \\
        \bottomrule
    \end{tabularx}
\end{table*}

\subsection{Ablation Study}
We conduct ablation studies on both the BIRD-CRITIC Open and PG subsets to analyze the contribution of key components in our framework. By selectively removing or simplifying individual components, we examine how the planning structure and critical tools affect overall performance.

\subsubsection{Ablation on Planning Structure}
We conduct an ablation study to examine the role of structured planning in our SQL critic framework.
As the full method allows the model to generate a tree-structured refinement plan and supports branching and rollback during execution process of the plan, we introduce a linear variant that restricts the refinement process to a single sequential path. Specifically, the LLM is only allowed to calling tools and generating SQLs step-by-step in this variant.
This design allows us to isolate the effect of tree-structured planning.

Removing the tree-structured planning mechanism and constraining the refinement process to a linear, step-by-step correction results in a substantial performance drop on both BIRD-CRITIC Open and PG subsets (Table~\ref{tab:ablation}). This performance degradation highlights the importance of structured planning with branching and rollback in SQL correction. Without the ability to explore alternative refinement paths, the linear variant is forced to commit to a single correction hypothesis at each step. Once an incorrect intermediate decision is made, subsequent refinements are restricted to this flawed trajectory, leading to error accumulation or premature convergence to suboptimal corrections. In contrast, the tree-structured plan enables the model to defer commitment, recover from invalid intermediate decisions, and maintain robustness against early-stage mistakes.

\subsubsection{Ablation on Key Tools}
We further perform ablation studies on the key tools integrated into our framework to evaluate their individual contributions during refinement.
Specifically, we remove the SQL execution tool (including run SQL and inspect columns) and the user query ambiguity detection tool, respectively.
These ablations aim to quantify the importance of execution-based feedback and ambiguity handling in guiding reliable SQL correction, with results summarized in Table~\ref{tab:ablation}.

As shown in Table~\ref{tab:ablation}, removing either the SQL execution tool or the user query ambiguity detection tool consistently degrades performance on both the BIRD-CRITIC Open and PG subsets, confirming the importance of both execution-based feedback and explicit ambiguity handling for reliable SQL correction. In particular, removing the SQL execution tool results in a substantially larger performance drop. This suggests that execution feedback plays a critical role during refinement by providing an interface for the LLM to inspect data distributions or intermediate results. Without such an interface, the evidence available for branch selection and rollback becomes significantly weaker, preventing the tree-structured debugging process from fully exercising its advantages.
In contrast, the ambiguity detection tool primarily affects early-stage user intent disambiguation. Its removal increases the likelihood that the refinement process proceeds along correction paths based on misinterpreted user intent, which in turn leads to degraded performance.

\subsection{Evaluation on Real-world Data}
\label{sec:tls_result}
In this section, we evaluate \model on a real-world industrial dataset and compare it with representative baselines, aiming to assess its effectiveness and efficiency in practical deployment scenarios.

\subsubsection{Evaluation Settings}
We describe the real-world dataset, task formulation, and evaluation criteria used in our experiments below.

\ipara{Real-world Dataset - TLS.}
We evaluate \model on a real-world Text-to-SQL dataset collected from an industrial log analytics service, referred to as \textbf{TLS}. TLS is a SQL-like domain-specific query language for large-scale log analysis, which differs substantially from mainstream SQL dialects such as PostgreSQL and SQLite\footnote{TLS syntax documentation: \url{https://www.volcengine.com/docs/6470/1335024?lang=zh}, \url{https://www.volcengine.com/docs/6470/73638?lang=zh}}.
Due to its non-standard syntax and domain specificity, TLS is not covered by existing Text-to-SQL benchmarks and has limited publicly available training data.
The dataset consists of real user-issued natural language queries and their corresponding ground-truth TLS queries executed in the production system. It spans 288 TLS topics (tables), with an average of 21.33 columns per table, reflecting substantial schema diversity. Unlike SQL correction benchmarks, the TLS dataset does \emph{not} provide erroneous SQL as input; instead, each method must directly generate a TLS query from the user request, making this a pure Text-to-TLS generation task.

\begin{figure*}[t]
    \centering
    \includegraphics[width=0.74\textwidth]{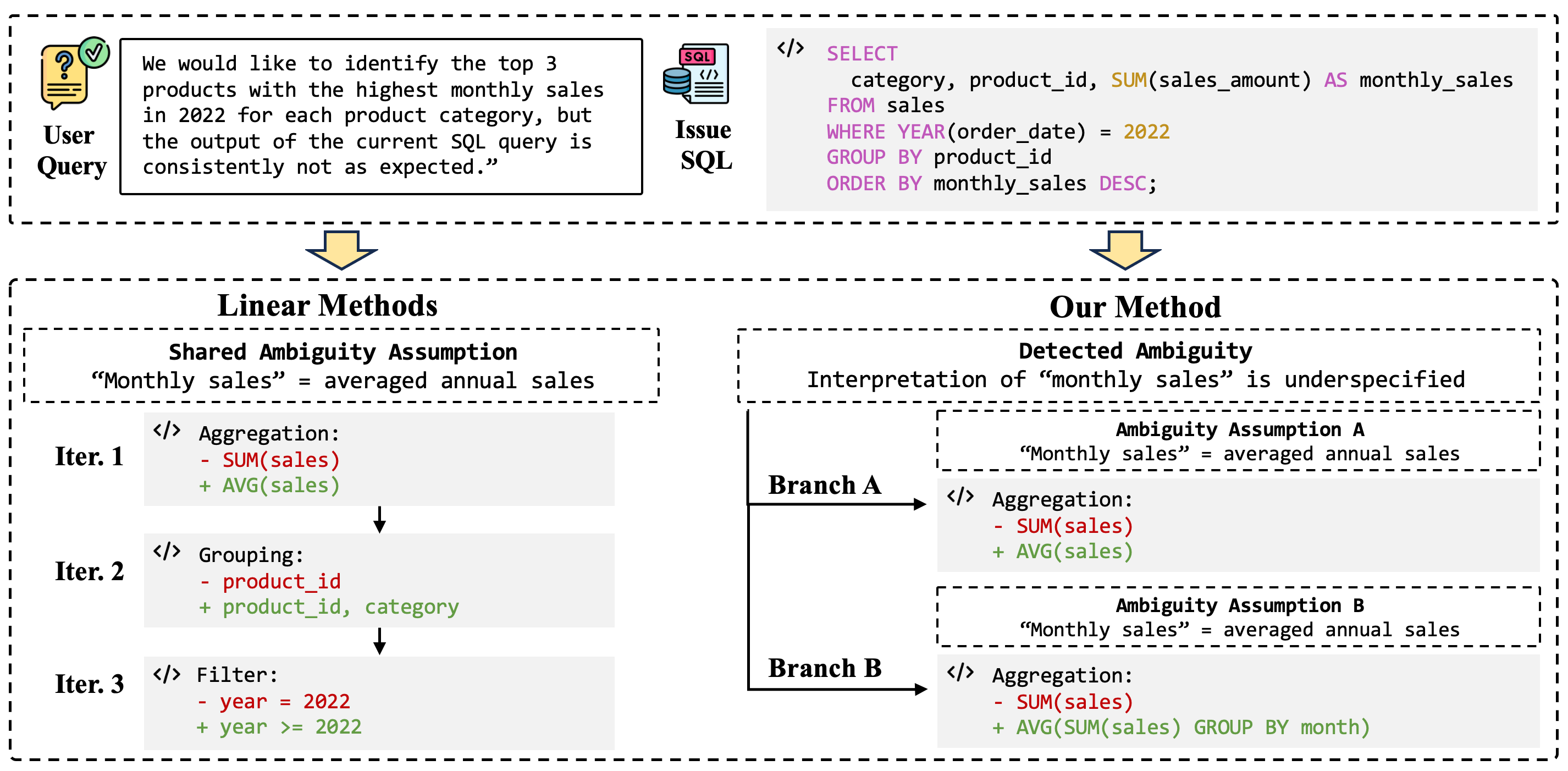}
	\caption{\label{fig:exp_case} Comparison between linear refinement and our tree-structured debugging on a representative SQL correction case. Linear refinement method iteratively applies local fixes under a fixed but incorrect semantic assumption, leading to repeated failure. \model branches on ambiguous interpretations, validates each branch independently, and backtracks from invalid paths, successfully correcting the SQL.}
\end{figure*}

\ipara{Task Definition and Evaluation Criteria.}
Given a natural language user query, each method first generates an initial TLS query, which is then executed against the target log system.
A prediction is considered correct if its execution result matches that of the ground-truth TLS query.
When critic-based or agent-based methods are applied, they iteratively diagnose and refine the initially generated TLS query based on execution feedback, rather than correcting a pre-existing erroneous query.
We report both the initial accuracy of the generated TLS queries and the final accuracy after refinement.
In addition, we measure the end-to-end latency introduced by different critic modules, covering the complete refinement process, to assess their practical usability in real-world systems.

\ipara{Baselines.}
We evaluate multiple Text-to-TLS generation backbones, including strong general-purpose LLMs and fine-tuned Text-to-SQL/Text-to-TLS models.
To study the impact of different correction strategies, we further equip these backbones with various critic mechanisms, including LLM-based critics and representative agent-based SQL correction methods.
Notably, the agent-based baselines used in this setting are consistent with those evaluated on BIRD-CRITIC, allowing us to examine their effectiveness when transferred to a real-world Text-to-TLS scenario.
This design enables a controlled comparison of different critic modules under the same generation and execution conditions.

\subsubsection{Results on Real-world Data}
We further analyze the effectiveness of \model on the TLS dataset, which represents a real-world Text-to-TLS task collected from an industrial log analytics system. Experimental results are shown in table~\ref{tab:tls_results}, we have the following findings:

\textbf{Effectiveness.}
\model consistently achieves the highest accuracy on the TLS dataset across all evaluated settings.
Compared with the initial Text-to-TLS generation, the introduction of our critic mechanism leads to significant performance improvements, indicating that execution-aware refinement is highly effective in real-world scenarios.
In addition, our approach outperforms existing LLM-based critics and agent-based baselines, demonstrating its stronger ability to identify and correct semantic and logical issues in generated TLS queries.
These results confirm that \model can substantially enhance end-to-end Text-to-TLS performance on realistic user queries.

\textbf{Efficiency.}
We further report the runtime latency of different critic-based methods to assess their practical usability.
Although our approach performs iterative refinement, its additional latency remains moderate and comparable to other agent-based baselines.
Importantly, the observed performance gains are achieved without introducing prohibitive execution overhead, suggesting that \model strikes a favorable balance between effectiveness and efficiency.
This makes it suitable for deployment in real production systems where both accuracy and response time are critical.

\begin{figure*}[t]
    \centering
    \includegraphics[width=0.95\textwidth]{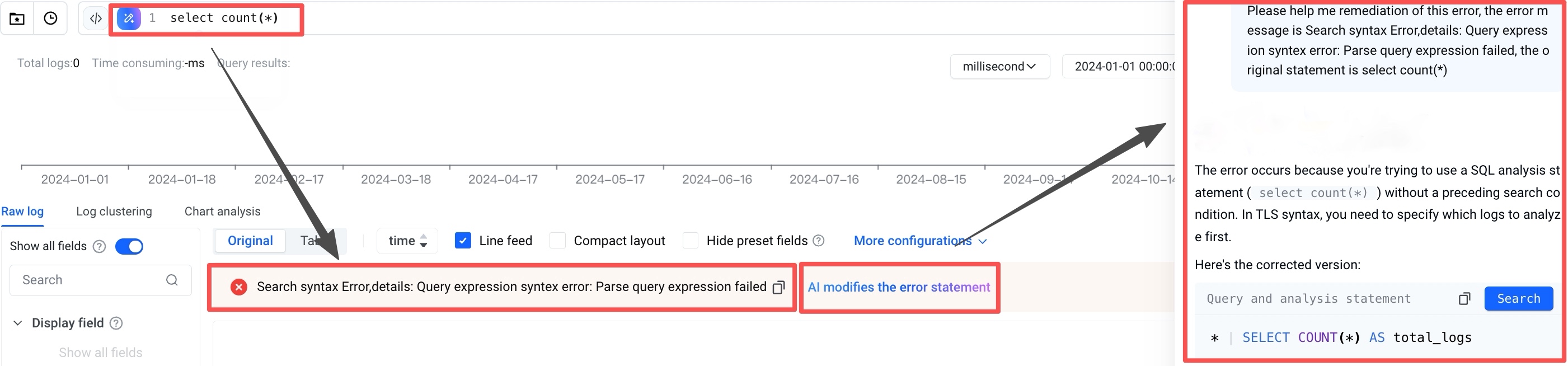}
	\caption{\label{fig:product} Screenshot of the Text-to-TLS API interface and error TLS repair copilot in the production environment.}
\end{figure*}

\textbf{Robustness.}
The effectiveness of \model is consistently observed across multiple LLM backbones, indicating that the proposed refinement mechanism does not rely on a specific model architecture or training condition.
Moreover, \model remains effective in the pure Text-to-TLS setting, where queries are generated from scratch rather than corrected from pre-existing erroneous inputs.
These results suggest that our approach is robust to both backbone variations and task settings, and can reliably improve generation quality in realistic scenarios where no explicit error signals are provided upfront.

\subsubsection{Discussion}
The results on the TLS dataset indicate that our critic-based framework is both effective and practical for real-world Text-to-TLS systems, supporting its deployment in industrial pipelines.

During \textbf{training and data construction}, LLM-generated queries are commonly used for dataset synthesis and augmentation, where uncorrected errors directly affect data quality.
\model offers a training-free correction mechanism that consistently improves the correctness of generated queries across different backbones, making it suitable for automated data cleaning without additional annotation or fine-tuning costs.

During \textbf{online inference}, imperfect query generation remains unavoidable even with strong LLMs.
The observed accuracy gains and moderate latency overhead show that our critic can be integrated as a post-generation refinement module, improving system reliability while satisfying practical efficiency requirements.
Moreover, its stable performance across multiple backbones suggests that it can function as a general-purpose correction layer, rather than being tied to a specific model.

\subsection{Case Study}

We present a representative real-world case to illustrate how our tree-structured debugging framework resolves semantic ambiguities that systematically trap linear refinement-based agents.
The user query asks for the top-3 products with the highest monthly sales in 2022 for each product category.
While seemingly simple, the phrase ``monthly sales'' is underspecified and admits multiple plausible interpretations, making it a common failure mode for SQL correction systems.

\subsubsection{Failure of Linear Refinement}
As shown in Figure~\ref{fig:exp_case} (left), the base model generates an initial SQL query that aggregates sales over the entire year 2022 and directly ranks products by the aggregated value.
This reflects an implicit assumption that ``monthly sales'' can be derived from annual aggregation, without explicitly modeling month-level granularity.

Linear refinement-based agents inherit this early semantic commitment.
Although execution feedback indicates incorrect results, the agent never revisits the interpretation of ``monthly sales''.
Instead, it repeatedly applies local modifications—such as changing aggregation functions, adjusting grouping keys, or relaxing temporal filters—while preserving the same flawed assumption.

Despite modifying different SQL components across iterations, all refinements remain constrained to the same annual-aggregation view.
As a result, the agent repeatedly produces syntactically valid but semantically incorrect SQL and fails to recover after multiple execution-feedback cycles.

\subsubsection{Tree-Structured Debugging with Backtracking}
In contrast, \model explicitly identifies ``monthly sales'' as an ambiguous semantic decision point and branches the debugging process accordingly, as shown in Figure~\ref{fig:exp_case} (right).

The agent explores multiple alternative hypotheses in parallel.
One branch follows the annual-aggregation interpretation adopted by linear methods, while another introduces explicit month-level aggregation before computing product-level statistics.
Each branch is independently validated through SQL execution.

When the annual-aggregation branch produces inconsistent results, the system backtracks to the ambiguity node and continues exploration along the alternative path.
By validating and pruning semantic hypotheses early, the agent successfully converges to the correct interpretation and generates the correct SQL.
This branching-and-backtracking mechanism enables recovery from erroneous early commitments that fundamentally limit linear refinement approaches.

\subsection{Cost Analysis}
We analyze the runtime cost of \model on a real-world business benchmark, where we measure the number of LLM API calls, end-to-end latency, and token consumption. Our evaluation includes both reasoning and non-reasoning LLMs. For token accounting, we only report the visible output tokens returned by the API, excluding internal reasoning tokens that are not exposed to users, which ensures a consistent and practical measurement across different model types.

Overall, \model invokes the LLM API 4.97 times per instance on average, with a mean visible token consumption of 8,984.5 tokens (measured with GPT-o3). The API calls mainly arise from the central agent’s decision-making steps, LLM-based tools such as user query ambiguity detection, and iterative SQL generation during refinement. Despite the use of multiple tools, the total number of interaction rounds remains moderate, reflecting a compact workflow design.

In terms of latency, the average end-to-end runtime is around 3 minutes when using non-reasoning models such as GPT-5, and increases to approximately 4–5 minutes with reasoning models such as GPT-o3. This latency difference is primarily attributed to the model-side inference time of reasoning models, rather than additional API calls or tool interactions introduced by our framework. SQL execution and validation also contribute to the overall runtime but account for a relatively small portion compared to LLM inference. These results indicate that the higher latency observed with reasoning models is largely a property of the models themselves, while our workflow does not introduce substantial additional overhead.

\subsection{Application}
\model has been integrated into the Text2TLS interface of Volcengine Log Service and is currently deployed in production to support real-world log analysis queries. In particular, the TLS repair capability shown in Figure~\ref{fig:product} incorporates our proposed tree-structured debugging framework to automatically correct erroneous TLS queries during execution. Its online effectiveness is reflected by the results on the Text2TLS benchmark reported in Section~\ref{sec:tls_result}.
\section{Conclusion}
In this paper, we present a training-free SQL correction framework that formulates debugging as a tree-structured, plan-guided process, enabling systematic exploration of alternative repair strategies and recovery from incorrect decisions. By integrating lightweight diagnostic tools for ambiguity analysis, execution feedback, and clause-level validation, our approach improves the reliability of SQL correction without requiring additional supervision or model fine-tuning.

We evaluate our method on the public BIRD-Critic benchmark and observe state-of-the-art performance across multiple SQL dialects. Beyond offline benchmarks, we deploy the framework in a real-world industrial Text-to-SQL pipeline, where it significantly improves end-to-end execution accuracy with practical latency and cost. The results demonstrate that our approach is robust across different LLM backbones and suitable for production use, supporting both online SQL generation and large-scale training data curation. Overall, this work shows that structured, plan-based SQL correction provides an effective and deployable solution for improving SQL correctness in real-world LLM-based systems.



\bibliographystyle{ACM-Reference-Format}
\bibliography{sample}

\end{document}